\documentclass[runningheads]{llncs}

\usepackage{accv}

\usepackage{accvabbrv}

\usepackage{graphicx}

\usepackage[accsupp]{axessibility}  

\usepackage{hyperref}

\usepackage{orcidlink}
\usepackage{multirow}
\usepackage{amsmath}
\usepackage{amssymb}
\usepackage{booktabs}
\usepackage[most]{tcolorbox}

\begin{document}

\title{OTT3R: Multi-View 3D Reconstruction and Fast Dataset Generation at 1\% Compute} 

\titlerunning{OTT3R: 3D Reconstruction at 1\% Compute}

\author{Brandon Leblanc\orcidlink{0009-0009-9590-0119} \and
Charalambos Poullis\orcidlink{0000-0001-5666-5026}}

\authorrunning{B.~Leblanc and C.~Poullis}

\institute{Immersive and Creative Technologies Lab, Concordia University, Montreal, Canada\\
\email{brandon.leblanc@mail.concordia.ca, charalambos.poullis@concordia.ca}}

\maketitle

\begin{figure}[h]
    \centering
    \newcommand{\teaserpanel}[1]{%
        \begin{tcolorbox}[
            width=0.32\linewidth, height=0.24\linewidth,
            boxrule=0.5pt, arc=4pt, left=0pt, right=0pt, top=0pt, bottom=0pt,
            colframe=black, colback=white, valign=center, halign=center,
            nobeforeafter, box align=top]
        \includegraphics[width=\linewidth,height=\linewidth,keepaspectratio]{#1}
        \end{tcolorbox}}
    \teaserpanel{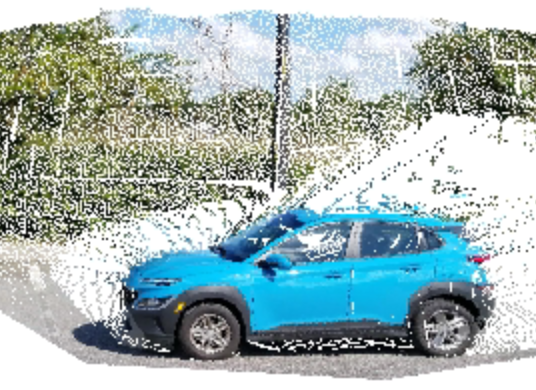}\hfill
    \teaserpanel{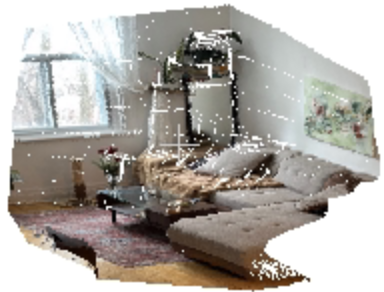}\hfill
    \teaserpanel{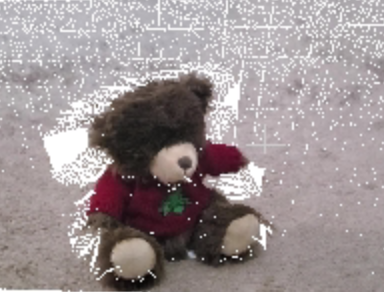}
    \caption{Sample 3D reconstructions produced by OTT3R on diverse scenes: outdoor object, indoor room, and object-centric capture, all from RGB-only input.}
    \label{fig:teaser}

\end{figure}

\begin{abstract}
Feed-forward 3D reconstruction models have achieved impressive performance by scaling model and dataset size, but their cost excludes most research groups and precludes edge deployment. Additionally, generating 3D supervision without sensors still relies on slow, unreliable Structure-from-Motion, as the community lacks a COLMAP-like system for neural 3D pseudo-label generation. We present OTT3R (RGB-Only Tiny Transformer for 3D Reconstruction), a knowledge distillation framework that addresses both problems on a single workstation equipped with 2 GPUs. Distilling $\pi^3$ (959M parameters) into a 102M-parameter student yields 9.4$\times$ compression and up to 7$\times$ faster inference, trained at 1.6\% of VGGT's training compute. An integrated pseudo-label pipeline offers a reliable, high-throughput alternative to COLMAP, generating dense per-pixel point maps and SE(3) camera poses for a 667K-image corpus in 3.5 hours on two commodity GPUs and succeeding on every sequence we tested, including those where COLMAP fails. The general student tracks the teacher on in-distribution monocular depth and, zero-shot, outperforms COLMAP on 7-Scenes and on DTU completion, but it does not replace the teacher on out-of-distribution multi-view geometry. The deployable artifact is the domain-specialized student: after specialization at 0.2\% compute, it is 4$\times$ more accurate than COLMAP on 7-Scenes at 980$\times$ throughput, with near-teacher completion. Code is available at \url{https://github.com/TheFourthKaramazov/OTT3R}.

\keywords{feed-forward 3D reconstruction \and knowledge distillation \and pseudo-label generation \and efficient deep learning}
\end{abstract}

\section{Introduction}
\label{sec:intro}

Feed-forward 3D reconstruction has advanced rapidly since DUSt3R~\cite{wang2024dust3r} demonstrated that dense geometry and implicit camera poses could be regressed directly from image pairs without following the classical Structure-from-Motion (SfM) and Multi-View Stereo pipeline~\cite{schonberger2016sfm, hartley2004multiple, snavely2006photo, frahm2010building, schonberger2016pixelwise}. Subsequent work extended this paradigm with models like Fast3R~\cite{yang2025fast3r}, VGGT~\cite{wang2025vggt}, and $\pi^3$~\cite{wang2026pi3}, achieving state-of-the-art results by processing all views jointly through global attention mechanisms, thus avoiding the cost associated with post-hoc alignment of pairwise predictions and subsequent pose recovery via PnP~\cite{lepetit2009epnp}. These advances follow a familiar trajectory in deep learning: rapid gains driven by exponential increases in model capacity and dataset scale~\cite{kaplan2020scaling, hoffmann2022training, dosovitskiy2021image}. Fast3R trains on 128 A100 GPUs for 6 days with approximately $650\mathrm{M}$ parameters; $\pi^3$ contains $959\mathrm{M}$ parameters and requires similar infrastructure; more recent work, VGGT-$\Omega$~\cite{wang2026vggt}, scales the model size by 10$\times$, the number of scenes by 15$\times$, and trains on 128 H100 GPUs. Although these models produce high-quality reconstructions, their size and computational requirements create two limitations: they cannot be deployed on the edge devices required by applications such as robotics and computer-assisted surgery where 3D reconstruction is most valuable~\cite{wu2022tinyvit, baruch2021arkitscenes, 2026distill3r, letellier2025foundry}, and they cannot be trained or adapted by researchers without access to industrial-scale compute. In this work we ask how much feed-forward 3D reconstruction is achievable at roughly 1\% of the training compute used by these systems, on commodity hardware that a single research group or practitioner can realistically own.

This accessibility problem has led the research community to approaches that avoid the core training challenge entirely. Most recent work targets inference-only acceleration through token merging, sparse attention, quantization~\cite{shu2025litevggt, fastvggt_arxiv, wang2025fastervggtblocksparseglobal, chen2025co, quantvggt_arxiv}, or partial fine-tuning~\cite{ren2025fin3r}: useful for deployment, but unable to adapt models to new domains or let researchers iterate on architectures. The few methods that do train new architectures still require datacenter infrastructure~\cite{ren2026speed3r, chen2026reliev3r}, leaving most groups unable to participate. To our knowledge, only Distill3R~\cite{2026distill3r} has demonstrated single-workstation feed-forward 3D distillation, but its student is effectively restricted to monocular reconstruction; we extend that line of work to multi-view reconstruction, configurable sampling, parallelizable pseudo-label generation, and a 3$\times$ more storage-efficient pipeline.

A related problem compounds this accessibility gap: generating 3D supervision for new datasets remains difficult. Traditional SfM pipelines \cite{schonberger2016sfm} are slow, processing at approximately 0.3 FPS in our evaluation experiments at reduced resolution, and require hours per scene at the higher resolutions needed to avoid failure. They are unreliable, with documented failure modes including textureless regions, repetitive structures, and dynamic objects that produce insufficient point matches; failure rates increase further at lower resolutions practitioners often use to keep runtime tractable. They are fragile, requiring careful parameter tuning for each scene type. As a result, researchers are increasingly turning to feed-forward models like $\pi^3$ and VGGT to generate pseudo-labels for downstream tasks~\cite{peddi2026towards}, but the community lacks a COLMAP-like system for neural 3D pseudo-label generation. In particular, VGGT-$\Omega$ demonstrates that feed-forward 3D models continue to benefit from larger datasets, extending the trend established from DUSt3R to VGGT, yet the community lacks the internet-scale 3D supervision required to sustain this scaling because such data is prohibitively costly to generate and annotate~\cite{gu2024conceptgraphs}. OTT3R's pseudo-label pipeline requires only unposed RGB images and generates a full suite of dense 3D supervision for hundreds of thousands of images in \textit{hours} rather than the days or weeks COLMAP would require, while remaining reliable on sequences where SfM fails.

To address both problems, we introduce OTT3R (\cref{fig:teaser}), a knowledge distillation framework that compresses $\pi^3$ into a compact student model while providing infrastructure for domain-specific training using neurally-generated pseudo-labels in place of SfM, all at between 0.2\% and 1.6\% of the training compute used by large feed-forward 3D models. We extend single-workstation distillation with 2 GPUs to multi-view reconstruction by leveraging the teacher's explicit SE(3) outputs for direct pose supervision and a configurable sampling system that produces the wide-baseline pairs pose supervision requires. At this budget, the general student tracks the teacher on in-distribution depth and, zero-shot, beats COLMAP on 7-Scenes and DTU completion, but does not replace the teacher out of distribution; the deployable artifact is the domain-specialized student. Our main contributions are as follows:
\begin{enumerate}
\item A pseudo-label generation pipeline that serves as a reliable, high-throughput neural alternative to COLMAP, featuring configurable overlap sampling strategies, manifest-based dataset management for easy extensibility, multi-sample generation from long sequences for data augmentation, quality filtering via confidence and depth-edge masking, and overlap-based sampling for dataset compression and large-scale generation. The pipeline generates dense supervision signals, including camera poses, for a 667K-image corpus in 3.5 hours on two commodity GPUs, and produces a cache roughly $3\times$ smaller per image than the closest prior distillation work~\cite{2026distill3r}.

\item A 102M-parameter student architecture that achieves 9.4$\times$ compression from the 959M-parameter teacher, with up to 7$\times$ faster inference and peak memory requirements reduced by 2--4$\times$. Full general training completes on 2$\times$ RTX 6000 Ada GPUs at approximately 1.6\% of VGGT's training compute~\cite{wang2025vggt} (64 A100s for 9 days), and domain specialization on 7-Scenes~\cite{shotton2013scene} runs in 23 hours at roughly 0.2\%. Zero-shot, the general student beats COLMAP with dense MVS on 7-Scenes and on DTU completion; the domain-specialized student is 4$\times$ more accurate on 7-Scenes at 980$\times$ the throughput.

\item The first distillation framework with direct SE(3) pose supervision via teacher outputs, enabling multi-view reconstruction, complemented by a hybrid confidence loss for calibrated uncertainty. A systematic analysis of design choices at this compute budget shows that pose supervision and sampling diversity produce categorical improvements, while loss formulation, encoder backbone, and model size within reasonable bounds produce only small, robust differences; no choice we ablated closes the remaining gap to the teacher, which we attribute to the compute budget.
\end{enumerate}

\section{Related Work}
\label{sec:related}
Below we provide a brief overview of relevant work grouped according to their paradigm.

\textbf{From Procedural SfM to Feed-Forward 3D Reconstruction.} Recovering 3D geometry from unposed images has historically relied on Structure-from-Motion (SfM) \cite{schonberger2016sfm, hartley2004multiple, snavely2006photo, agarwal2009building, frahm2010building}, a multi-stage procedure combining local features \cite{detone2018superpoint, dusmanu2019d2net, zhao2023aliked, yi2016lift, lowe2004distinctive}, pairwise matching \cite{sarlin2020superglue, sun2021loftr, edstedt2024roma, lindenberger2023lightglue_iccv}, geometric verification, and bundle adjustment, often followed by dense multi-view stereo \cite{yao2018mvsnet, gu2020cascade, schonberger2016pixelwise, furukawa2015multi}. Its sequential nature compounds failure modes such as weak textures, repeated structure, dynamic objects and runtime grows superlinearly with image count \cite{wu2013towards, ozyesil2017survey, oliensis2000critique}. Learned matching \cite{sarlin2020superglue, edstedt2023dkm, jin2021image}, self-supervised descriptors \cite{tyszkiewicz2020disk}, and differentiable bundle adjustment \cite{wang2024vggsfm, tang2018ba, wei2020deepsfm, teed2018deepv2d} close part of the gap, but the system remains fragile outside curated benchmarks and slow at the resolutions needed to avoid failure. OTT3R replaces this workflow with a feed-forward neural pipeline that produces 3D supervision.

Feed-forward networks predict a dense per-pixel point map that fuses pose and geometry into a single output and is learned end-to-end from large curated 3D datasets \cite{reizenstein2021co3d, dai2017scannet, baruch2021arkitscenes, savva2019habitat, li2018megadepth, yao2020blendedmvs, yeshwanth2023scannetpp, deitke2023objaverse}. The DUSt3R formulation is pairwise: any multi-view reconstruction requires a post-hoc alignment step using pose solvers \cite{lepetit2009epnp, wang2023posediffusion, kendall2015posenet} whose cost grows rapidly with view count. Subsequent work introduces matching heads \cite{leroy2024mast3r}, sequential memory \cite{wang2025spann3r}, dynamic-scene handling \cite{zhang2025monst3r}, conditioning on priors \cite{jang2025pow3r}, sparse-view camera and appearance estimation \cite{zhang2025flare}, and multi-view stereo coupling \cite{cabon2025must3r, tang2024mvdust3r}. Fast3R \cite{yang2025fast3r}, VGGT \cite{wang2025vggt}, and $\pi^3$ \cite{wang2026pi3} replace pairwise reasoning with joint global attention over all views, producing scene-consistent reconstructions in one forward pass. These networks also predict per-pixel confidence \cite{kendall2017what, kendall2016modelling}, a signal for both supervision filtering and loss weighting. Their datacenter-scale training that consists of up to 128 H100 GPUs for VGGT-$\Omega$ \cite{wang2026vggt}, a dataset consisting of 4M sequences, and up to 10B parameter counts rule out edge deployment and place training beyond most research groups.

\textbf{Efficient Feed-Forward 3D.} Most efficiency work targets inference rather than training: token-level methods reduce work per forward pass via merging, sparse attention, or post-training quantization~\cite{fastvggt_arxiv, shu2025litevggt, chen2025co, wang2025flashvggt, wang2025fastervggtblocksparseglobal, sun2025avggtrethinkingglobalattention, quantvggt_arxiv, stary2025understandingmultiviewtransformers}, and memory or retrieval extensions handle longer sequences~\cite{yuan2026infinitevggt, zou2026attention}. Methods that do train new architectures, such as Speed3R~\cite{ren2026speed3r} and Reliev3R~\cite{chen2026reliev3r}, require 8 H20 GPUs and 64 NPUs respectively, sitting outside the commodity-hardware regime. These inference-time methods are orthogonal and complementary to our goal rather than baselines for it: they reduce the cost of an already-trained network while preserving its accuracy, whereas OTT3R targets training from scratch at $\sim$1\% compute. Because token merging, sparse attention, and quantization operate on a trained transformer at inference, they apply to the OTT3R student exactly as they do to the teacher, and can be stacked on top of it for further speedups.

\textbf{Knowledge Distillation.} Knowledge distillation \cite{hinton2015distillation, gou2020knowledge, touvron2021training} is the natural mechanism for training compact models under tight compute budgets. The DUNE encoder \cite{sariyildiz2025dune}, which we adopt as our backbone, shows that heterogeneous co-distillation can yield compact 3D-aware representations that exceed their teachers on downstream geometric tasks. Prior 3D distillation efforts target neighboring problems such as monocular depth \cite{wu2023adudepth, song2023msdpt} and multi-view depth, where KD-MVS \cite{ding2022kdmvs} distills depth without ground-truth labels but assumes known camera parameters and does not pursue compression. The per-scene effort of \cite{dutt2024multiview} distills DUSt3R one scene at a time and does not produce a general model. Closest to our work, Distill3R \cite{2026distill3r} demonstrated that a feed-forward 3D model (Fast3R) can be compressed into a 72M-parameter student trainable on a single workstation with 2 GPUs using offline teacher caching inspired by TinyViT \cite{wu2022tinyvit} together with a confidence-weighted geometric loss. Distill3R has two principal limitations: the absence of camera-pose supervision and a temporally-local sampling regime that preserves correspondence consistency but rarely exposes the student to viewpoint changes large enough to learn pose estimation, yielding effectively monocular reconstructions. OTT3R targets both, using the teacher's explicit SE(3) outputs for direct pose supervision and a configurable sampler that produces the wide-baseline transitions that learning pose estimation requires. 

\section{Method}
\label{sec:method}

\subsection{Problem Formulation}
Given $N$ uncalibrated RGB images $\mathcal{I} = \{I_1, \dots, I_N\}$ depicting a scene, we aim to train a compact student model $\mathcal{S}$ that approximates the outputs of a large teacher model $\mathcal{T}$ ($\pi^3$, 959M parameters). For each image $I_k$, both models produce a local point map $\mathbf{P}^{\ell}_k \in \mathbb{R}^{H \times W \times 3}$ of per-pixel 3D coordinates in the camera frame (denoted $\mathbf{P}^{s}_k$ and $\mathbf{P}^{t}_k$ for the student and teacher respectively in the losses below), a per-pixel confidence map $C_k \in \mathbb{R}^{H \times W}$, and a camera pose $\mathbf{T}_k \in SE(3)$ as a camera-to-world transformation. We do not store global point maps; global geometry is recovered when needed by transforming local points with the predicted poses, $\mathbf{P}^{g}_k = \mathbf{T}_k \cdot \tilde{\mathbf{P}}^{\ell}_k$, where $\tilde{\mathbf{P}}^{\ell}_k$ denotes homogeneous coordinates. This avoids redundant per-pixel supervision and reduces both storage and training cost. Additionally, \cite{wang2025vggt} observed that performance is better when using local point maps combined with predicted camera poses rather than directly predicting global point maps. Our objective is to minimize the discrepancy between student and teacher outputs through a hybrid distillation loss $\mathcal{L}_{\text{total}}$.

\subsection{Pseudo-Label Generation Pipeline}
\label{sec:pseudo}

\begin{figure}[t]
    \centering
    \includegraphics[width=\textwidth]{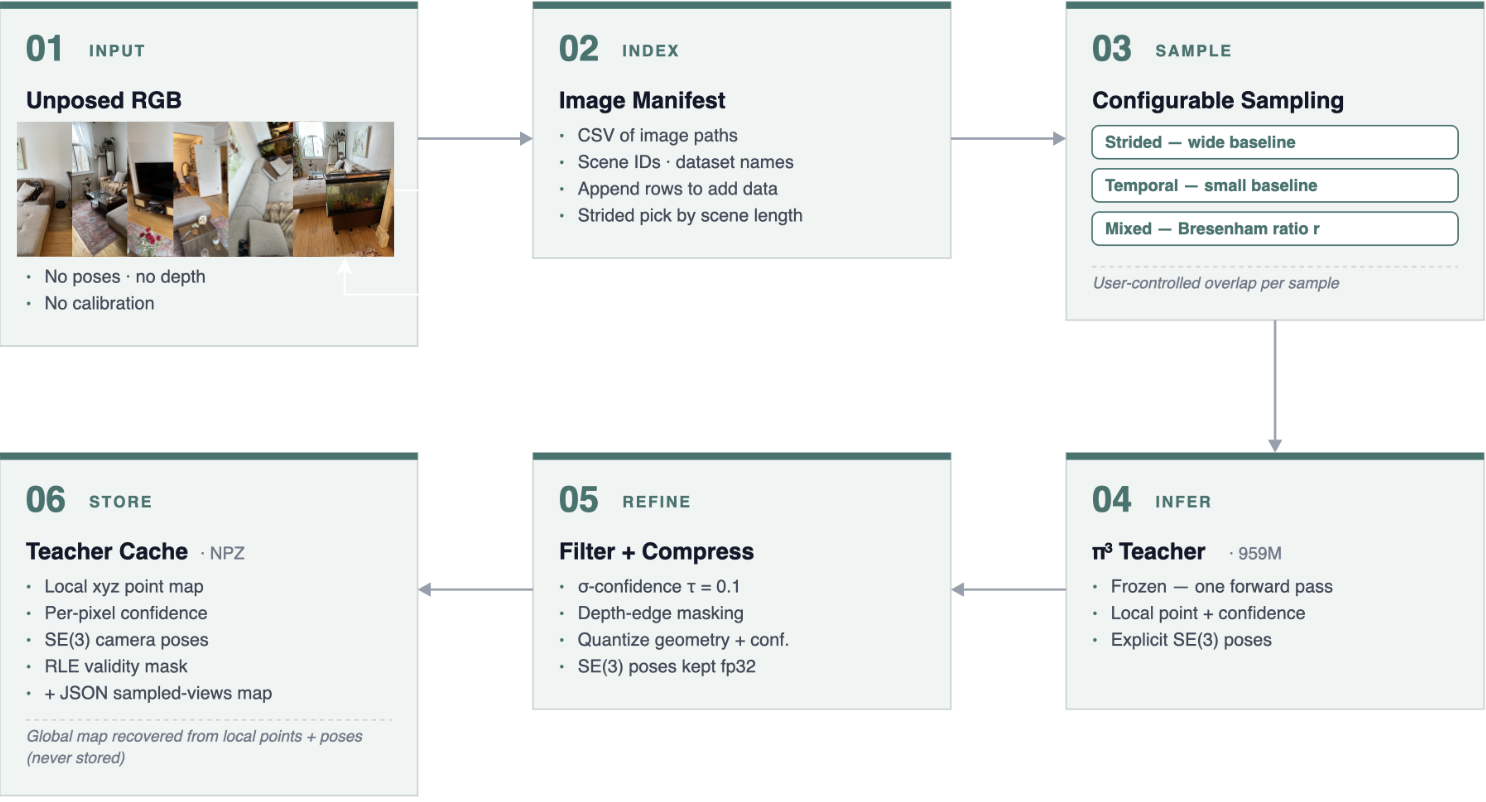}
    \caption{\textbf{Pseudo-label generation pipeline.} Given RGB images, we generate dense 3D supervision using $\pi^3$. Our configurable sampling strategy selects frames with either strided (hard) or temporal (easy) spacing. Teacher outputs are compressed and cached for efficient student training.}
    \label{fig:pipeline}
\end{figure}

We provide infrastructure for generating dense 3D supervision from RGB images at scale (\cref{fig:pipeline}). Unlike COLMAP, our pipeline requires no parameter tuning, is orders of magnitude faster, and did not fail on any sequence we tested, including the low-resolution and low-texture sequences where SfM breaks down.

\textbf{Dataset Distillation and Sampling Strategies.} 
To ensure training is feasible on limited hardware, we first compress and distill external datasets using strided sampling based on scene length. Because these datasets are organized temporally, we reduce their size while preserving geometric variance by scaling the sampling stride to scene length: a 1000-image sequence is heavily subsampled, while a 20-image scene is kept in full. This compresses a multi-terabyte corpus into a manageable 400 GB dataset. Training on temporally-adjacent frames alone produces students that handle easy transitions but fail on challenging viewpoint changes. Our pipeline provides two base sampling strategies and a configurable mix of them. \textit{Strided sampling} spreads views across the full sequence by computing base positions via linear interpolation, then adding random jitter to introduce variation while maintaining coverage. Indices are clipped rather than wrapped to avoid artificial discontinuities, creating wide-baseline pairs that require the student to learn robust pose estimation. We show that the student model fails to learn pose estimation without strided sampling (see~\cref{sec:ablation}). \textit{Temporal sampling} selects consecutive frames with a small baseline for video and SLAM applications. \textit{Mixed sampling} combines both at a user-specified ratio using Bresenham-style interleaving: for sample index $i$ and hard ratio $r$, strided sampling is used when $\lfloor(i+1) \cdot r\rfloor > \lfloor i \cdot r\rfloor$, producing evenly distributed hard and easy samples regardless of the ratio.

\textbf{Quality Filtering, Storage, and Extensibility.} Before caching, each pixel undergoes quality filtering to ensure reliable supervision. We apply a sigmoid and threshold to discard low-confidence pixels because the teacher outputs raw confidence logits. We additionally discard depth-edge pixels, where adjacent relative depth changes sharply, as the teacher's predictions at these boundaries often contain interpolation artifacts. The final validity mask combines both criteria, and invalid pixels are excluded from all training losses. We compress teacher outputs before caching, quantizing geometry and confidence while keeping camera poses at full precision, which keeps storage and I/O manageable at scale. The pipeline uses a manifest-based architecture in which datasets are defined via CSV files listing image paths, scene IDs, and dataset names, so adding a new dataset requires only appending rows to the manifest.

\textbf{Scalability.} Cache generation and training are fully decoupled. Users generate pseudo-labels from many datasets (CO3D \cite{reizenstein2021co3d}, ARKitScenes \cite{baruch2021arkitscenes}, MegaDepth \cite{li2018megadepth}, etc.) in a single pass, then train on a subset without regenerating (the training setup is illustrated in Appendix~A): for example, filtering to indoor scenes alone for a domain-specific model. The pipeline generates dense 3D supervision for our full 667K-image corpus in 3.5 hours on two RTX 6000 Ada GPUs, and a 26K-image domain-specific cache in under 10 minutes. This is possible because generation is GPU-bound: at $\sim$23 images per second per GPU the $\pi^3$ forward pass dominates, so throughput is relatively stable across worker counts. Data loading is not a significant bottleneck on a single workstation, although this may change if the number of GPUs is scaled significantly. The pipeline can adapt to any number of GPUs and wall-clock scales linearly with GPU count (\cref{tab:cache_scaling}). It is both faster and more storage-efficient than the closest prior distillation pipeline: Distill3R generates 450K images in 11.3 hours ($\sim$11 img/s) and caches them in $\sim$600\,GB~\cite{2026distill3r}, whereas we generate at more than twice the per-GPU throughput and store 667K images in 283\,GB, a 3$\times$ improvement in per-image cost achieved by recovering global geometry from local points and poses rather than caching it explicitly.

\subsection{Student Architecture}
We initialize the encoder with DUNE ViT-Small~\cite{sariyildiz2025dune}, shared across all $N$ views. DUNE's heterogeneous co-distillation from 2D and 3D teachers provides a strong initialization for dense 3D regression, which we adapt to the teacher's output distribution; we observe comparable performance using DINOv2~\cite{oquab2024dinov2} (see~\cref{sec:ablation}), indicating the framework is robust to the choice of encoder within this family. A multi-layer transformer uses an alternating attention pattern for cross-view reasoning, following the global-attention designs of VGGT~\cite{wang2025vggt} and $\pi^3$~\cite{wang2026pi3}. Three parallel transformer decoders produce the final outputs from shared decoder features. The point decoder produces local 3D coordinates in each camera's reference frame, the confidence decoder outputs per-pixel certainty, and the camera decoder regresses SE(3) poses as camera-to-world transformations.

\subsection{Distillation Loss}
\label{sec:loss}

\textbf{Point Loss.} We use scale-invariant L1 loss with optimal scale alignment:
\begin{equation}
\mathcal{L}_{\text{point}} = \frac{1}{\sum_k |\mathcal{M}_k|} \sum_{k} \sum_{(i,j) \in \mathcal{M}_k} w_{k,ij} \cdot \| s^* \cdot \mathbf{p}^{s}_{k,ij} - \mathbf{p}^{t}_{k,ij} \|_1
\end{equation}

A single optimal scale $s^*$ per sample, shared across all $N$ views, is computed via weighted L1 alignment using the ROE solver \cite{wang2025moge, wang2026pi3} and detached from the gradient graph to prevent degenerate solutions. Depth-based weights $w = 1/(d + \epsilon)$ prioritize closer points where depth estimates are more reliable. The valid masks $\mathcal{M}_k$ are the sets of valid pixel coordinates $(i,j)$ in view $k$, defined by teacher confidence, preventing the student from destabilizing training by predicting low confidence on difficult regions or high confidence in early training when predictions are inaccurate.

\textbf{Camera Loss.} We directly supervise SE(3) poses, but compute them against teacher poses over all $\binom{N}{2}$ view pairs, providing dense supervision that captures both small and large viewpoint changes:
\begin{equation}
\mathcal{L}_{\text{camera}} = \frac{1}{\binom{N}{2}}\sum_{a<b} \left( \alpha \cdot \mathcal{L}_{\text{trans}}(a,b) + \mathcal{L}_{\text{rot}}(a,b) \right)
\end{equation}
where the sum is over all view pairs $(a,b)$, $\mathcal{L}_{\text{trans}}(a,b) = \mathrm{Huber}_{\delta}(s^* \mathbf{t}^s_{ab} - \mathbf{t}^t_{ab})$ is Huber loss on relative translation, using the same detached per-sample $s^*$ as the point loss so that the scale-free student is compared in the teacher's scale, and $\mathcal{L}_{\text{rot}}$ is geodesic angle loss on relative rotation: $\mathcal{L}_{\text{rot}}(a,b) = \arccos\left(\frac{\text{tr}(\mathbf{R}^s_{ab}{}^\top \mathbf{R}^t_{ab}) - 1}{2}\right)$. The weighting $\alpha=100$, adopted from $\pi^3$, balances gradient magnitudes between rotation and translation.

\textbf{Hybrid Confidence Loss.} Regressing student confidence directly to teacher values via L1 collapses to teacher mimicry, while using only BCE with binary labels saturates the confidence head and ignores the teacher's uncertainty estimates (see~\cref{sec:ablation}). Our hybrid formulation blends both per pixel:
\begin{equation}
\mathcal{L}_{\text{conf}} = c^t \cdot \mathcal{L}_{\text{BCE}} + (1 - c^t) \cdot \mathcal{L}_{\text{L1}}
\end{equation}
where $c^t$ is the teacher confidence at each pixel. The BCE labels are derived from the student's own reconstruction error: $y = \mathbf{1}[\bar{e} < \theta]$ where $\bar{e}$ is the depth-weighted mean point error. In high-confidence regions ($c^t \approx 1$), the loss teaches the student to predict confidence based on its geometry quality. In low-confidence regions ($c^t \approx 0$), the loss becomes L1 regression to the teacher's estimate, deferring to teacher uncertainty rather than learning unreliable self-assessment in difficult areas. Although the BCE labels depend on the student's current reconstruction error and therefore evolve during training, we observe stable convergence in practice; we compare the hybrid against pure-L1 and pure-BCE variants in~\cref{sec:ablation}.

\textbf{Normal Loss.} We additionally enforce surface consistency with a normal loss $\mathcal{L}_{\text{normal}}$ \cite{wang2026pi3}, computing per-pixel normals from cross-products of adjacent point differences and minimizing the angular difference between predicted and teacher normals. This term is added to the point loss to ensure that geometric and surface-orientation supervision are optimized together. The total loss is: 
\begin{equation}
\mathcal{L}_{\text{total}} = \underbrace{\mathcal{L}_{\text{point}} + \lambda_{\text{normal}} \mathcal{L}_{\text{normal}}}_{\mathcal{L}_{\text{point}}^{+}} + \lambda_{\text{cam}} \mathcal{L}_{\text{camera}} + \lambda_{\text{conf}} \mathcal{L}_{\text{conf}}
\end{equation}
where $\mathcal{L}_{\text{point}}^{+}$ denotes the point loss with the normal term included.

\section{Experiments}
\label{sec:experiments}

\subsection{Implementation Details}

\textbf{Model Architecture.} Our student uses a DUNE ViT-Small \cite{sariyildiz2025dune} encoder with 384-dim, 12 layers, 6 heads, patch size of 14, and 21M parameters. The main decoder has 512-dim, 12 layers, and 8 heads. Task decoders for points, confidence, and camera each use 4 transformer layers with 512-dim embeddings and 8 heads. The camera head outputs 384-dim features. The total parameters are 102M, a 9.4$\times$ compression from the 959M-parameter teacher.

\textbf{Training.} We train on 2 RTX 6000 Ada GPUs (182 BF16 TFLOPS each at FP32 accumulate) using PyTorch with bf16-mixed precision. Measured in TFLOPS$\cdot$h at this precision, VGGT~\cite{wang2025vggt} costs $64 \times 312 \times 216\,\mathrm{h} \approx 4.31\mathrm{M}$ (312 TFLOPS per A100); our general student costs $2 \times 182 \times 192\,\mathrm{h} \approx 69.9\mathrm{K}$ (1.62\%), full pseudo-label generation 3.5\,h $\approx$ 1.3K (0.03\%), and the 7-Scenes student 23\,h $\approx$ 8.4K (0.19\%). Because $\pi^3$ initializes from VGGT and trains further on 64 A100s, these ratios are upper bounds relative to our teacher; as is standard in distillation, we exclude the frozen public $\pi^3$ and DUNE checkpoints. Batch size is 2 per GPU with 4$\times$ gradient accumulation, giving an effective batch of 16, for 200 epochs on 667K images (33K 20-view samples) drawn from CO3D~\cite{reizenstein2021co3d}, MegaDepth~\cite{li2018megadepth}, ARKitScenes~\cite{baruch2021arkitscenes}, ScanNet++~\cite{yeshwanth2023scannetpp}, Habitat~\cite{savva2019habitat}, and BlendedMVS~\cite{yao2020blendedmvs}. We use AdamW with learning rate $10^{-4}$, weight decay 0.01, and cosine decay to $10^{-5}$. Full training completes in approximately 8 days. For 7-Scenes specialization we train from scratch on pseudo-labels generated from the 7-Scenes training split, for 500 epochs with minimum learning rate $10^{-6}$, completing in approximately 23 hours. Quality filtering uses confidence threshold $\tau = 0.1$ on the sigmoid teacher output and a 3\% relative-depth threshold for depth-edge detection. Strided sampling jitters bin positions by $\pm 35\%$. Loss settings: Huber threshold $\delta = 0.1$ for translation, BCE error threshold $\theta = 0.02$ for the hybrid confidence loss, and loss weights $\lambda_{\text{cam}} = 0.1$, $\lambda_{\text{conf}} = 0.05$, $\lambda_{\text{normal}} = 1.0$. These loss settings follow $\pi^3$; a grid search over alternatives produced changes that were either negligible or harmful.

\subsection{Results}

\begin{table}[t]
\centering
\caption{Multi-view reconstruction on 7-Scenes~\cite{shotton2013scene} and DTU~\cite{aanaes2016large} (Acc/Comp in cm, $\pi^3$ protocol, 518$\times$224). The general student is evaluated zero-shot on both benchmarks; the domain student is specialized on 7-Scenes at $\sim$0.2\% of VGGT's training compute. COLMAP runs at 640$\times$480 and is evaluated only on the scenes it reconstructs.}
\label{tab:7scenes}

\resizebox{\linewidth}{!}{%
\begin{tabular}{lccccccc}
\toprule
& & \multicolumn{3}{c}{7-Scenes} & \multicolumn{3}{c}{DTU} \\
\cmidrule(lr){3-5} \cmidrule(lr){6-8}
Method & Params & Acc $\downarrow$ & Comp $\downarrow$ & FPS $\uparrow$ & Acc $\downarrow$ & Comp $\downarrow$ & FPS $\uparrow$ \\
\midrule
$\pi^3$ Teacher~\cite{wang2026pi3} & 959M & 1.52 & 2.03 & 40 & 0.11 & 0.13 & 59 \\
\midrule
COLMAP + dense MVS$^\ddagger$~\cite{schonberger2016sfm} & -- & 24.27 $\pm$ 1.20 & 57.56 $\pm$ 8.41 & 0.28 $\pm$ 0.02 & \textbf{0.68} & 6.69 & 0.32 \\
Distill3R$^\dagger$~\cite{2026distill3r} & 72M & 8.76 & 8.88 & 78 & -- & -- & -- \\
OTT3R (general, zero-shot) & 102M & 10.16 & 34.84 & 273 & 0.91 & \textbf{3.62} & \textbf{331} \\
\textbf{OTT3R (domain)} & 102M & \textbf{6.16} & \textbf{2.84} & \textbf{274} & n/a & n/a & n/a \\
\bottomrule
\end{tabular}%
}\\[2pt]
{\scriptsize\raggedright
$^\dagger$Per-view monocular evaluation~\cite[Sec.~IV-B, Tab.~II]{2026distill3r}; no multi-view result is available.
$^\ddagger$7-Scenes: mean $\pm$ std over 5 runs. Failure rate 9\% (7-Scenes) and 8\% (DTU) at 640$\times$480, 28\% and 64\% at 518$\times$224; failed scenes are excluded.\par}
\end{table}

\begin{table}[t]
\centering
\begin{minipage}[t]{0.55\linewidth}
\centering
\caption{Camera pose on CO3D \cite{reizenstein2021co3d}. The teacher is measured against ground truth~\cite{wang2026pi3}; student rows measure agreement with the teacher poses it distills, not absolute accuracy. Median translation in teacher scale.}
\label{tab:pose}
\resizebox{\linewidth}{!}{%
\begin{tabular}{llcccc}
\toprule
Method & Ref. & RRA@5 $\uparrow$ & RRA@30 $\uparrow$ & Med.\ rot.$^\circ$ $\downarrow$ & Med.\ trans.\ $\downarrow$ \\
\midrule
$\pi^3$ Teacher~\cite{wang2026pi3} & GT & -- & 99.0 & -- & -- \\
\midrule
OTT3R (temporal) & $\pi^3$ & 61.9 & 96.1 & 2.8  & 0.047 \\
OTT3R (strided)  & $\pi^3$ & 38.7 & 65.3 & 15.3 & 0.264 \\
OTT3R (all)      & $\pi^3$ & 50.3 & 80.7 & 4.9  & 0.081 \\
\bottomrule
\end{tabular}%
}
\end{minipage}\hfill
\begin{minipage}[t]{0.43\linewidth}
\centering
\caption{Student depth quality vs teacher on diverse datasets.}
\label{tab:student_depth}
\resizebox{\linewidth}{!}{%
\begin{tabular}{llccc}
\toprule
Dataset & Model & Params & Abs Rel $\downarrow$ & $\delta < 1.25$ $\uparrow$ \\
\midrule
\multirow{2}{*}{ARKitScenes~\cite{baruch2021arkitscenes}} & $\pi^3$ & 959M & 3.64\% & 98.8\% \\
& OTT3R (base) & 102M & 4.91\% & 97.0\% \\
\midrule
\multirow{2}{*}{Habitat~\cite{savva2019habitat}} & $\pi^3$ & 959M & 3.96\% & 98.3\% \\
& OTT3R (base) & 102M & 7.05\% & 93.9\% \\
\midrule
\multirow{3}{*}{7-Scenes~\cite{shotton2013scene}} & $\pi^3$ & 959M & 2.50\% & 98.9\% \\
& OTT3R (base) & 102M & 13.07\% & 85.7\% \\
& OTT3R$^\dagger$ & 102M & 5.72\% & 98.1\% \\
\bottomrule
\end{tabular}%
}\\[2pt]
{\scriptsize $^\dagger$Domain-specific student.}
\end{minipage}
\end{table}

\textbf{Multi-view Reconstruction.} We evaluate multi-view reconstruction on 7-Scenes~\cite{shotton2013scene}, a standard benchmark and the primary target of our domain-specific training, and on DTU~\cite{aanaes2016large}, a second benchmark outside our training data, following $\pi^3$'s official protocol with Sim3 (Umeyama) alignment, ICP refinement, and accuracy/completion metrics under normal consistency. On 7-Scenes the keyframe interval is 20 and the evaluation set is unseen during training. All evaluations are at 518$\times$224 resolution unless otherwise noted. For a fair dense-to-dense comparison, the COLMAP baseline runs the complete pipeline: feature extraction, exhaustive matching, sparse reconstruction, and dense multi-view stereo (MVS), so that it produces the same dense per-pixel geometry our metrics evaluate, not only a sparse point cloud. The failed scenes are not counted towards the COLMAP reconstruction metrics to ensure fairness.

\begin{figure}[t]
    \centering
    \begin{tabular}{@{}cccc@{}}
        \includegraphics[width=0.23\linewidth]{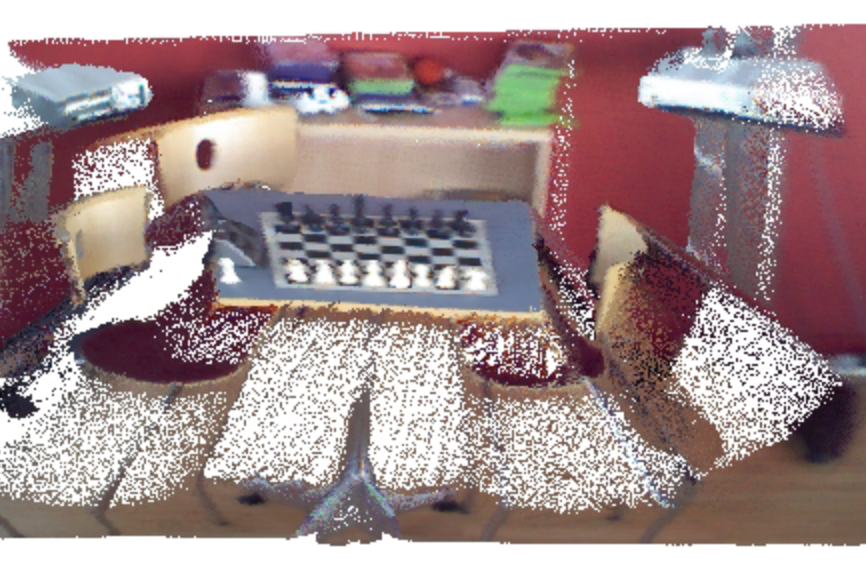} &
        \includegraphics[width=0.23\linewidth]{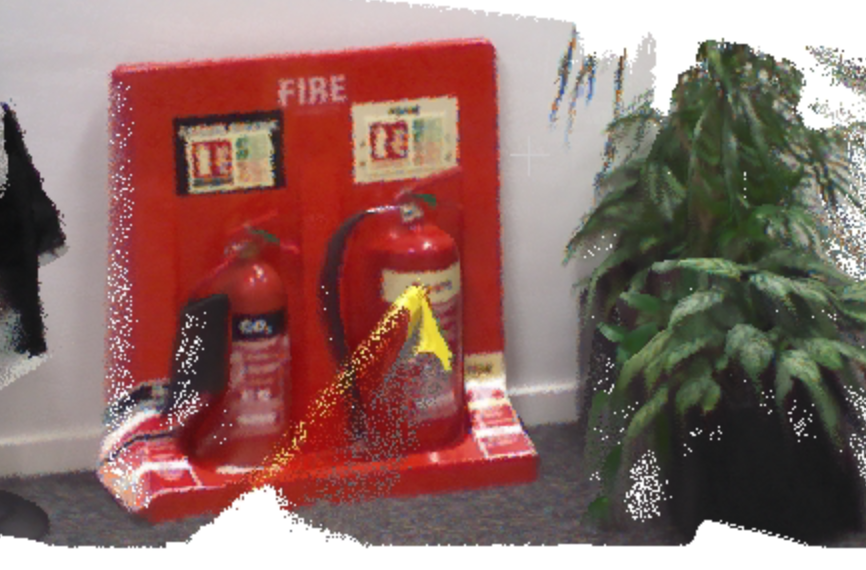} &
        \includegraphics[width=0.23\linewidth]{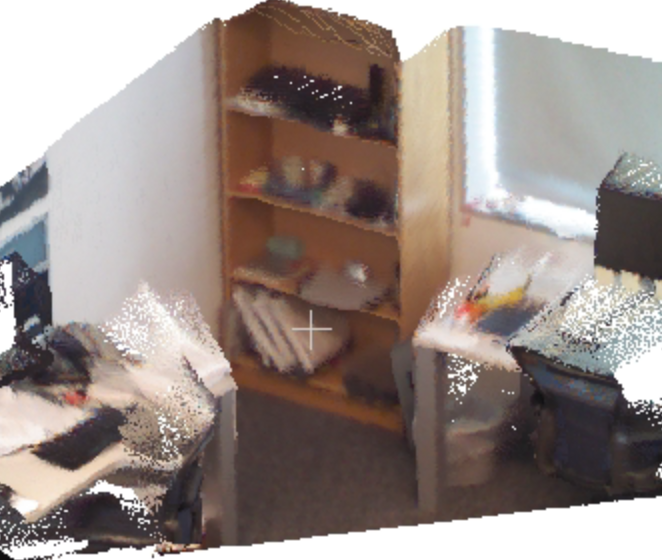} &
        \includegraphics[width=0.23\linewidth]{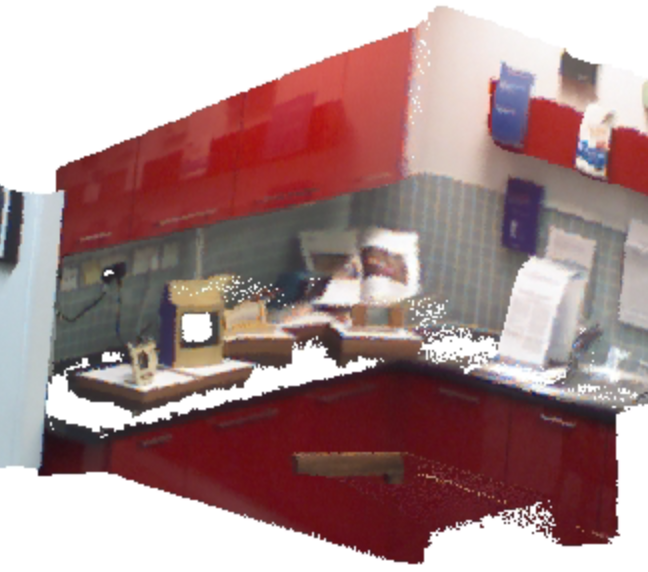} \\
        \includegraphics[width=0.23\linewidth]{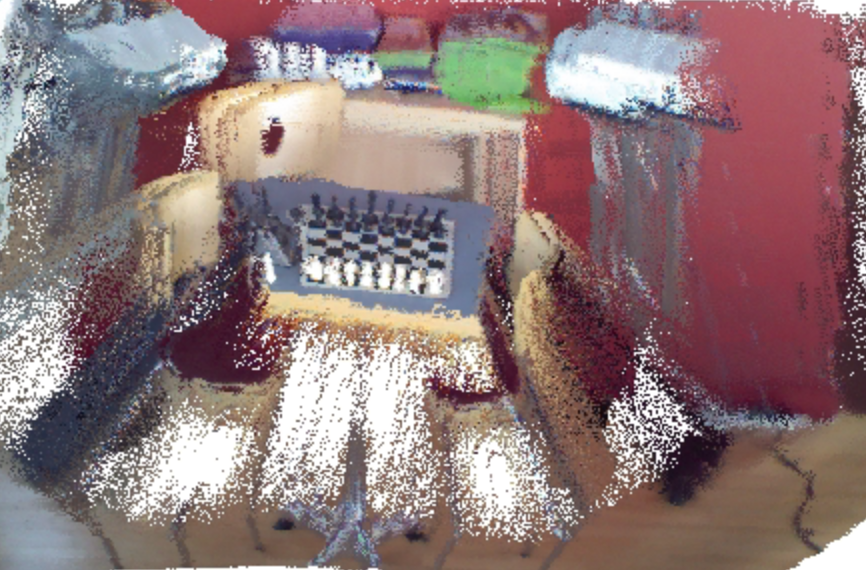} &
        \includegraphics[width=0.23\linewidth]{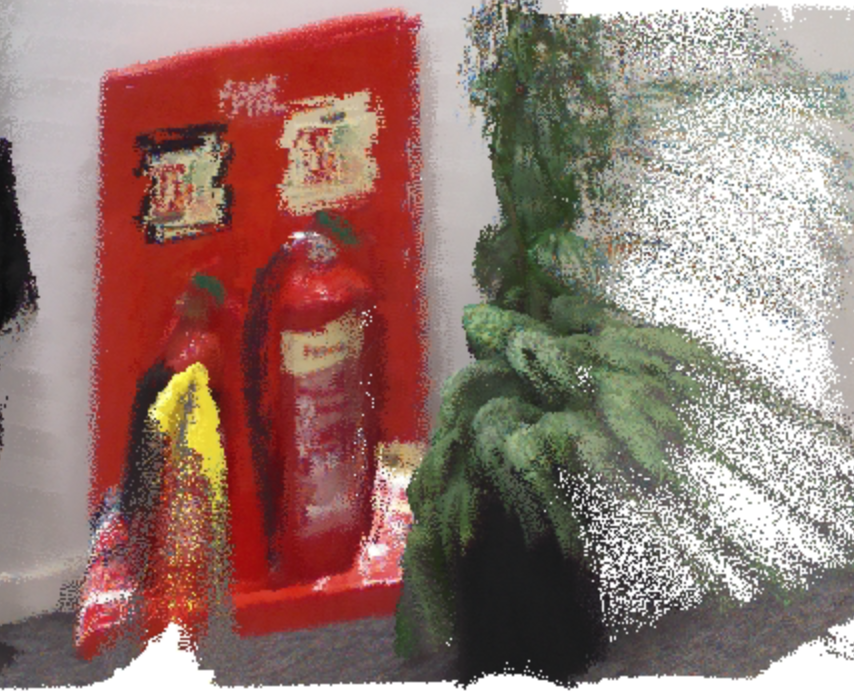} &
        \includegraphics[width=0.23\linewidth]{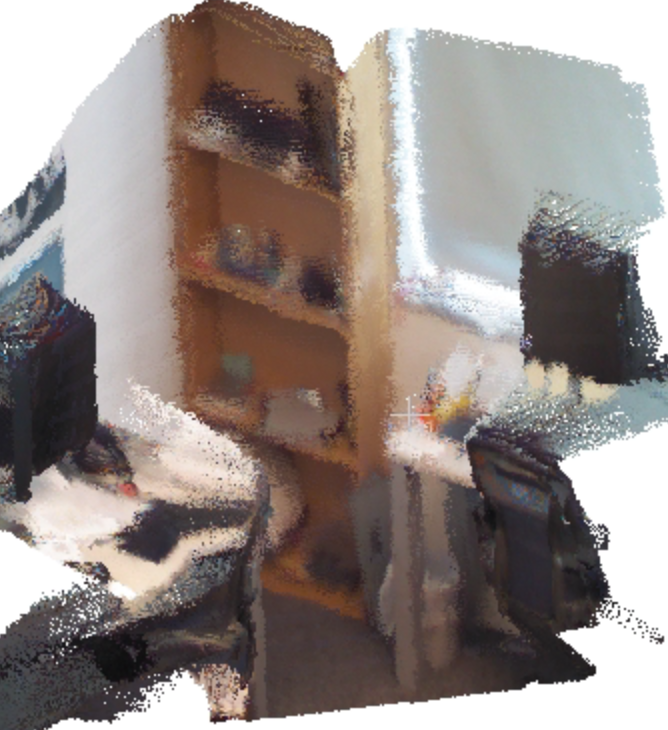} &
        \includegraphics[width=0.23\linewidth]{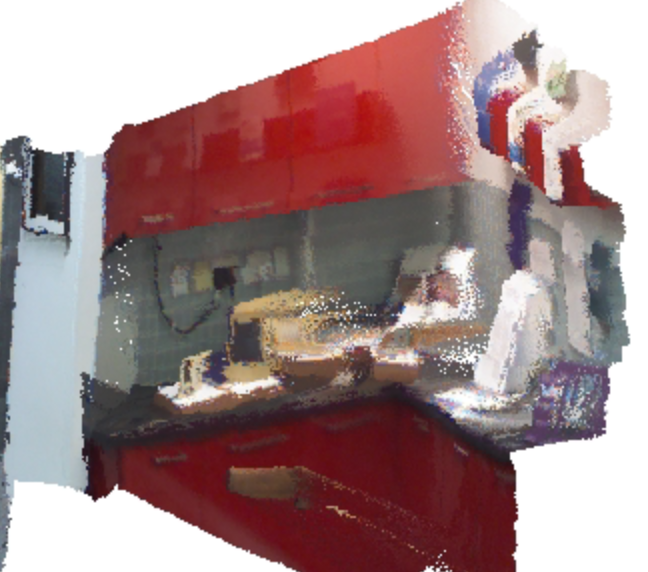} \\
        Chess & Fire & Office & Kitchen \\
    \end{tabular}
    \caption{Qualitative multi-view reconstruction on 7-Scenes test set. Top row: $\pi^3$ teacher (959M). Bottom row: OTT3R student (102M).}
    \label{fig:7scenes_qual}
\end{figure}

On 7-Scenes the headline comparison is against COLMAP, the SfM tool practitioners often use for room-scale reconstruction. Our domain-specific student is 4$\times$ more accurate than COLMAP and runs 980$\times$ faster, while reaching near-teacher completion at one-seventh the latency (\cref{tab:7scenes} and \cref{fig:7scenes_qual}). Zero-shot, the general student also beats COLMAP on 7-Scenes and on DTU completion but does not reach the teacher, as its out-of-distribution poses are too inaccurate to align views; specialization closes most of this gap. To make the comparison favorable to SfM, COLMAP is run at 640$\times$480 against our 518$\times$224. This is a real handicap for us, because COLMAP's success rate collapses at lower resolution, from 91\% to 72\% on 7-Scenes and from 92\% to 36\% on DTU. COLMAP entries are reported as the mean over 5 runs due to RANSAC non-determinism, with $\pm$1.20 cm accuracy and $\pm$8.41 cm completion variance across runs. Even with extra pixels and averaging across runs, COLMAP still performs worse on all 7-Scenes metrics and fails on roughly 9\% of 7-Scenes and 8\% of DTU sequences at the upscaled resolution, although it remains more accurate than the general student on DTU; we therefore position our pipeline on reliability, throughput, and completion rather than per-point accuracy.

\Cref{tab:7scenes} also includes Distill3R, the closest prior distillation work, with one important caveat: Distill3R has no pose head or loss, and its global point maps fail to align views~\cite[Sec.~V-C]{2026distill3r}; its numbers come from the strictly easier monocular setting~\cite[Sec.~IV-B, Tab.~II]{2026distill3r}, where each view is reconstructed independently and no global view alignment is required. Despite this asymmetry, OTT3R's multi-view metrics outperform Distill3R's monocular metrics by 30\% on accuracy, 68\% on completion, and 3.5$\times$ on throughput, reflecting the combination of direct SE(3) pose supervision, sampling diversity, and an improved pseudo-label pipeline. No compact multi-view model trained from scratch at comparable compute has been released, so the strongest learned baseline is the teacher itself; inference-time accelerations (\cref{sec:related}) reduce a trained model's FLOPs, not its training cost.

\textbf{Camera Pose Estimation.} The capability that distinguishes OTT3R from prior single-workstation distillation is direct SE(3) pose supervision, so we evaluate pose quality directly rather than only through reconstruction. We adopt the relative-pose protocol of~\cite{wang2026pi3}, comparing the relative rotation between every image pair within a sample; this is invariant to the global coordinate frame and to scene scale, and therefore requires no trajectory alignment. We evaluate on CO3D over 14{,}460 twenty-view samples ($\sim$5.5M pairs) and report relative rotation accuracy (RRA@5 and RRA@30, the fraction of pairs whose relative rotation lies within $5^\circ$ or $30^\circ$ of the reference) together with the median rotation error and the median relative translation error in the teacher's scale, using the median because the error distribution is heavy-tailed. The teacher fixes the ceiling: against ground truth, $\pi^3$ reaches 99.0 RRA@30 (and 88.4 AUC@30) on CO3D~\cite{wang2026pi3}, i.e.\ near-saturated pose accuracy. Because the teacher sits this close to ground truth, agreement with it is informative about absolute accuracy, though not a direct measure of it, and is precisely the quantity distillation optimizes. \Cref{tab:7scenes} also offers indirect ground-truth evidence: a single Sim3 aligns the fused cloud to ground-truth geometry, so near-teacher completion requires accurate relative poses.

\Cref{tab:pose} shows the student reproduces teacher pose closely on small-baseline transitions and trails on wide-baseline ones. On temporally-adjacent pairs it matches the teacher to a median $2.8^\circ$, with 61.9\% of relative rotations within $5^\circ$ and 96.1\% within $30^\circ$; on strided wide-baseline pairs the median rises to $15.3^\circ$ and agreement falls to 38.7\% within $5^\circ$ and 65.3\% within $30^\circ$ (combined: $4.9^\circ$, 50.3\%, 80.7\%). Median relative translation error follows the same pattern (0.047 temporal, 0.264 strided, 0.081 combined). Even the wide-baseline figure is far above chance: uniformly random rotations fall within $30^\circ$ under 1\% of the time so the student has clearly learned wide-baseline pose; its fidelity is simply lowest where the task is hardest. The residual tail on strided pairs is dominated by front/back confusions on near-symmetric CO3D objects viewed from opposing sides, a failure the teacher shares, which is why we report medians. Qualitatively, however, the rotation gap is not visually apparent. See Appendix~B for student reconstructions on temporal small-baseline and strided wide-baseline samples: despite the wide-baseline rotation error, the strided reconstructions remain visually comparable to the small-baseline ones and structurally coherent; Appendix~C shows cases where they are not.

\textbf{Student Depth Quality and Domain Specialization.} For monocular depth, per-view depth is extracted from the Z component of predicted camera-space points and aligned to ground truth by median scaling before computing Abs Rel and $\delta < 1.25$. \Cref{tab:student_depth} reports depth on three datasets. The $\pi^3$ rows also give the quality of our pseudo-labels: under 4\% Abs Rel and over 98\% $\delta<1.25$ on both LiDAR-captured ARKitScenes and synthetic Habitat. On ARKitScenes, the general student stays close to the teacher: 97.0\% $\delta<1.25$ against the teacher's 98.8\%, with the Abs Rel gap under 1.3\%. On Habitat, the student remains competitive at 93.9\% $\delta<1.25$ versus 98.3\% and 7.05\% Abs Rel versus 3.96\%. This is a small gap given $9\times$ less capacity, a significantly smaller training set, and roughly 1.6\% of the compute. The general student degrades on 7-Scenes, the most out-of-distribution of the three (85.7\% $\delta<1.25$, 13.1\% Abs Rel), reflecting room geometry and scale statistics underrepresented in its training mix.

The framework's response to this degradation is to train a domain-specific student rather than chase scale. On 7-Scenes, domain specialization recovers $\delta<1.25$ from 85.7\% to 98.1\%, within 0.8\% of the teacher's 98.9\%, and more than halves Abs Rel (13.1\% to 5.7\%), while running 6.8$\times$ faster than the teacher. The end-to-end cost is principally the sub-10-minute cache generation and 23-hour training run, both on the same single-workstation setup used for the general student. Any practitioner can specialize a compact 3D reconstruction model to their target domain in under a day, recovering near-teacher quality at a fraction of the inference cost and compute.

\textbf{Inference Efficiency.} \Cref{tab:efficiency} reports inference latency and peak memory across view counts. OTT3R is 6.2$\times$ faster than its teacher at 32 views using roughly a quarter of the memory. The gap widens with sequence length, where at 128 views VGGT runs out of memory (48GB VRAM) and the teacher needs 12.3~GB while OTT3R completes in 2.2~s using 5.2~GB. This is enough margin for deployment on consumer and edge hardware where 1B-parameter models are not viable.

\begin{table}[t]
\centering
\begin{minipage}[t]{0.40\linewidth}
\centering
\caption{Cache generation scaling on RTX 6000 Ada and wall-clock for the full 667K-image cache.}
\label{tab:cache_scaling}
\resizebox{\linewidth}{!}{%
\begin{tabular}{cccc}
\toprule
GPUs & Workers & Throughput (img/s) & Full cache \\
\midrule
1 & 1  & 23.1 & 7h\,44m \\
1 & 2  & 23.0 & 7h\,58m \\
2 & 1  & 22.9 & 3h\,50m \\
2 & 2  & 22.9 & 3h\,37m \\
2 & 16 & 22.5 & 3h\,20m \\
\bottomrule
\end{tabular}%
}
\end{minipage}\hfill
\begin{minipage}[t]{0.57\linewidth}
\centering
\caption{Inference latency (s) and peak memory (GB) on RTX 6000 Ada at 518$\times$224.}
\label{tab:efficiency}
\resizebox{\linewidth}{!}{%
\begin{tabular}{lccccccc}
\toprule
& & \multicolumn{3}{c}{Latency $\downarrow$} & \multicolumn{3}{c}{Memory $\downarrow$} \\
\cmidrule(lr){3-5} \cmidrule(lr){6-8}
Method & Params & N=32 & N=64 & N=128 & N=32 & N=64 & N=128 \\
\midrule
VGGT~\cite{wang2025vggt} & 1B & 2.28 & 6.40 & OOM & 33.98 & 38.41 & OOM \\
Fast3R~\cite{yang2025fast3r} & 650M & 1.14 & 3.26 & 10.11 & 12.11 & 21.11 & 44.36 \\
$\pi^3$ Teacher~\cite{wang2026pi3} & 959M & 1.49 & 4.30 & 12.82 & 7.07 & 8.99 & 12.34 \\
Distill3R~\cite{2026distill3r} & 72M & 0.41 & 1.02 & 2.69 & 9.97 & 21.80 & 31.90 \\
\textbf{OTT3R} & \textbf{102M} & \textbf{0.24} & \textbf{0.68} & \textbf{2.20} & \textbf{1.83} & \textbf{2.93} & \textbf{5.20} \\
\bottomrule
\end{tabular}%
}
\end{minipage}
\end{table}

\subsection{Ablations}
\label{sec:ablation}

\textbf{Pose Supervision and Sampling Diversity Are Necessary.} Removing direct SE(3) camera supervision (\cref{tab:abl_loss}) reproduces the failure mode of prior single-workstation distillation: completion degrades by $17\times$ (2.84 to 49.57 cm) as the student loses the ability to align views into a coherent scene. Restricting training to temporal-only sampling (\cref{tab:abl_loss}) produces a similar collapse: completion degrades by $9\times$ (2.84 to 25.22 cm) because the student never sees the wide-baseline pairs pose supervision requires. Each ablation isolates a pipeline contribution: pose supervision enables multi-view reconstruction, and configurable sampling produces the training distribution that pose supervision needs.

\textbf{Confidence Loss Formulation.} Pure-L1 and pure-BCE degrade 3--7\%  compared to the hybrid (\cref{tab:abl_loss}) but differ in calibration: pure-L1 collapses to teacher mimicry (correlation 0.88) and pure-BCE saturates to overconfidence (mean 0.97 vs teacher 0.70), while the hybrid stays calibrated (correlation 0.79). The geometric gap is small; the hybrid is preferable mainly for downstream tasks that consume confidence as a signal.

\textbf{Encoder Choice.} DUNE \cite{sariyildiz2025dune} ViT-S and DINOv2 \cite{oquab2024dinov2} ViT-S at matched parameter count produce comparable reconstruction quality (\cref{tab:abl_arch}), with DINOv2-S slightly ahead on accuracy (5.74 vs 6.16\,cm) but slightly behind on completion. DINOv3 \cite{simeoni2025dinov3} ViT-S underperforms both with our specific architecture due to patch size differences requiring interpolation. Scaling to DINOv2-B (280M) degrades quality substantially (9.66\,cm accuracy, 20.97 cm completion), indicating that at this compute budget, additional encoder capacity is not productively used. Halving the student to 54M also degrades reconstruction quality. The framework is encoder-robust within this group at the chosen scale; capacity matters, but more capacity is not better at fixed compute.

\begin{table}[t]
\centering
\begin{minipage}[t]{0.45\linewidth}
\centering
\caption{Loss and sampling ablations on 7-Scenes (domain-specific setting, 500 epochs). Pose supervision and sampling diversity are necessary.}
\label{tab:abl_loss}
\resizebox{\linewidth}{!}{%
\begin{tabular}{lccc}
\toprule
Variant & Acc $\downarrow$ & Comp $\downarrow$ & NC1 $\uparrow$ \\
\midrule
Baseline (hybrid) & \textbf{6.16} & \textbf{2.84} & 0.626 \\
\midrule
No camera loss & 8.88 & 49.57 & 0.531 \\
Temporal-only sampling & 11.52 & 25.22 & 0.565 \\
\midrule
L1-only conf & 6.43 & 3.03 & 0.626 \\
BCE-only conf & 6.33 & 2.97 & \textbf{0.630} \\
\bottomrule
\end{tabular}%
}
\end{minipage}\hfill
\begin{minipage}[t]{0.53\linewidth}
\centering
\caption{Architectural ablations (same setting as \cref{tab:abl_loss}). DUNE and DINOv2-S are competitive at matched capacity; halving to 54M degrades quality.}
\label{tab:abl_arch}
\resizebox{\linewidth}{!}{%
\begin{tabular}{lcccc}
\toprule
Variant & Params & Acc $\downarrow$ & Comp $\downarrow$ & NC1 $\uparrow$ \\
\midrule
Baseline (DUNE-S) & 102M & 6.16 & \textbf{2.84} & 0.626 \\
DINOv2-S & 102M & \textbf{5.74} & 3.16 & \textbf{0.628} \\
DINOv2-B & 280M & 9.66 & 20.97 & 0.560 \\
DINOv3-S & 102M & 6.94 & 5.53 & 0.613 \\
54M student & 54M & 6.62 & 3.19 & 0.624 \\
\bottomrule
\end{tabular}%
}
\end{minipage}
\end{table}

\textbf{Limitations.} OTT3R's student does not match teacher multi-view geometric precision, particularly on out-of-distribution scenes; its strength is in-distribution monocular depth, where it tracks the teacher closely, and domain specialization that recovers near-teacher quality on a target distribution in under a day. Practitioners requiring high-fidelity general reconstruction should use the teacher, while those targeting a specific domain or deployment scenario can specialize on commodity hardware. Pose is evaluated against the teacher, which is the student's effective ceiling, rather than ground truth. The pipeline inherits the teacher's limitations on strongly dynamic or non-rigid scenes. We did not sweep the filtering thresholds, individually ablate confidence filtering, depth-edge masking, cache quantization, or the normal loss, or run multiple seeds. Representative failure cases appear in Appendix~C.

\section{Conclusion}
\label{sec:conclusion}

We presented OTT3R, a knowledge distillation framework that compresses feed-forward 3D reconstruction models into a 102M-parameter student trainable on a single workstation at roughly 1.6\% of the compute used by comparable feed-forward 3D models, together with a pseudo-label generation pipeline for downstream supervision. The framework's two practical artifacts are the student and the pseudo-label generation pipeline. The student runs at 274\,FPS, remains competitive with the teacher on in-distribution monocular depth, and after domain specialization at $\sim$0.2\% of the teacher's compute beats COLMAP by 4$\times$ on 7-Scenes multi-view accuracy. The pseudo-label generation pipeline can produce dense per-pixel supervision for hundreds of thousands of images in hours, scaling linearly with GPU count, providing a mechanism for generating the large-scale 3D supervision required for scaling 3D vision applications in areas such as robotics.

\subsubsection*{Acknowledgments.} This research was undertaken, in part, based on support from the Natural Sciences and Engineering Research Council of Canada Grant RGPIN-2021-03479 (NSERC DG) and through the Natural Sciences and Engineering Research Council of Canada CGS-M scholarship (NSERC CGS-M).

\bibliographystyle{splncs04}
\bibliography{main}

\appendix
\renewcommand{\theHsection}{appendix.\Alph{section}}
\renewcommand{\theHfigure}{appendix.\arabic{figure}}

\clearpage
\begin{center}
  {\Large\bfseries Supplementary Material}
\end{center}
\vspace{1em}

This supplementary material provides additional qualitative results and implementation details omitted from the main paper. Appendix~\ref{supp:train} details the offline training pipeline, Appendix~\ref{supp:qual} shows qualitative multi-view reconstructions under temporal and strided sampling, and Appendix~\ref{supp:fail} shows representative failure cases.

\section{Training Pipeline}
\label{supp:train}

\begin{figure}[h]
    \centering
    \includegraphics[width=\linewidth]{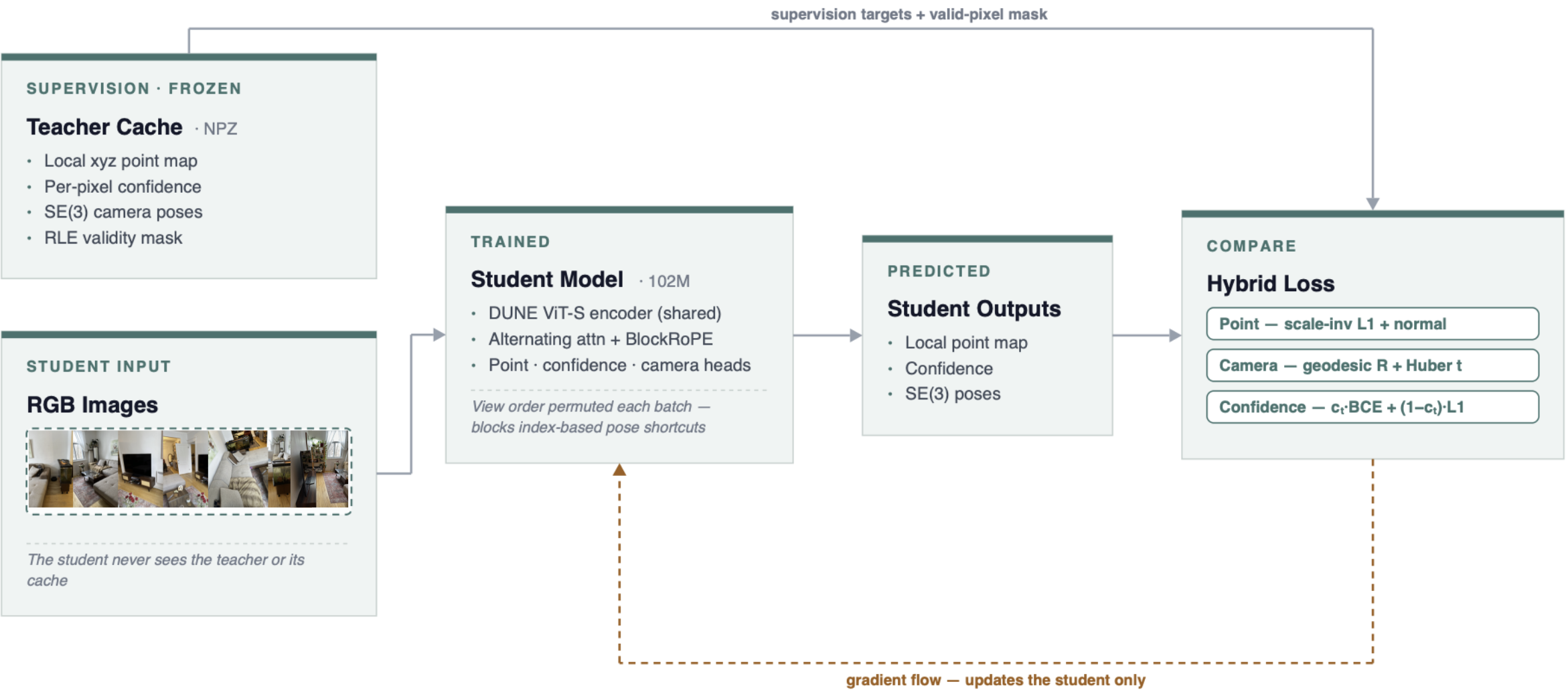}
    \caption{\textbf{Training pipeline.} The teacher is run once to
    populate an offline cache of quantized point maps, confidence, and
    SE(3) poses; the student then trains against this cache without
    re-invoking the teacher. Gradients update only the student.}
    \label{fig:supp_train}
\end{figure}

\cref{fig:supp_train} illustrates the offline distillation setup. Cache generation and student training are fully decoupled: the teacher $\pi^3$ is run once over all source datasets and its outputs are quantized and stored (main Section~3.2), after which training reads only from this cache and the teacher is never invoked again. This is what makes single-workstation training tractable: the 959M-parameter teacher forward pass, the dominant cost, is amortized across every epoch rather than repeated.

During training, each batch loads a pre-sampled group of $N$ views with their cached teacher point maps, confidence maps, and SE(3) poses. Gradients flow only through the student; the cached targets are constant. The student's three heads are supervised against the corresponding cached signals through the hybrid loss of main Section~3.4: the point and normal terms against teacher geometry, the camera term against teacher relative poses over all $\binom{N}{2}$ view pairs, and the confidence term against the student's own reconstruction error blended with teacher confidence. Because supervision is fully precomputed, the only online computation per step is the student forward and backward pass, so memory and wall-clock are governed by the 102M student alone rather than the teacher.

\section{Qualitative Multi-View Reconstructions}
\label{supp:qual}

We visualize OTT3R reconstructions on held-out CO3D sequences to illustrate the effect of sampling strategy on reconstruction quality. In each pair the same scene is reconstructed from temporal sampling (\emph{left}, consecutive frames with a small baseline) and strided sampling (\emph{right}, evenly-spaced frames spanning a wide baseline). Both regimes produce coherent geometry: despite the larger viewpoint changes and higher relative-rotation error of the strided case (main Section~4.2), its reconstructions remain visually comparable to the small-baseline ones, with the residual pose error manifesting as noise rather than visible structural misalignment in these examples; Appendix~\ref{supp:fail} shows failure cases. All reconstructions use 20 input views at $518\times224$ resolution. In every row, \emph{left} is temporal and \emph{right} is strided.

\par\medskip
\newcommand{\qualrow}[1]{%
  \noindent
  \includegraphics[width=0.49\linewidth]{figures/CO3D_sup/#1_temp.png}\hfill
  \includegraphics[width=0.49\linewidth]{figures/CO3D_sup/#1_stride.png}\\[12pt]%
}

\qualrow{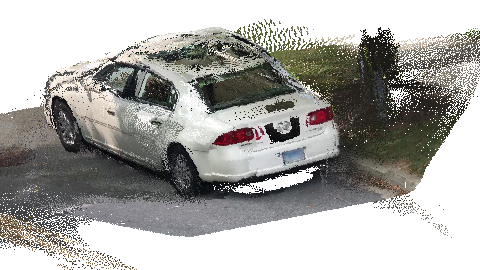}
\qualrow{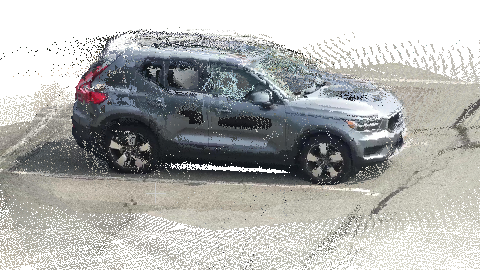}
\qualrow{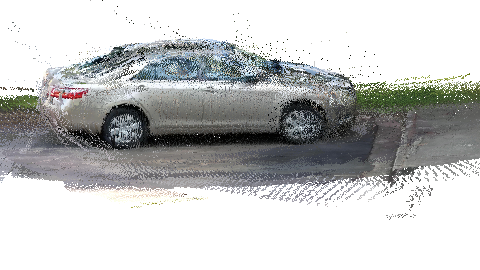}
\qualrow{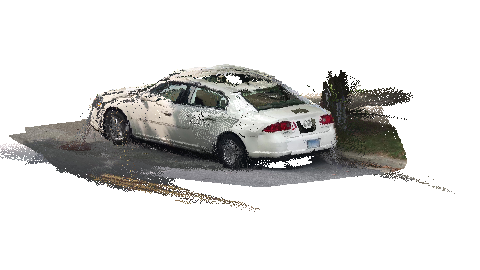}
\qualrow{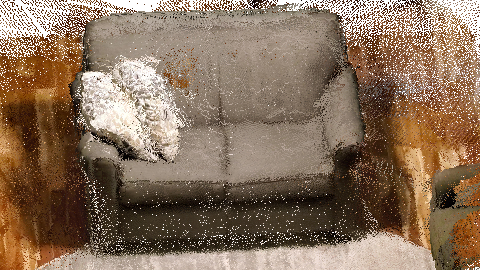}
\qualrow{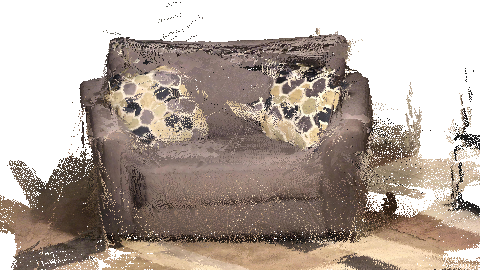}
\qualrow{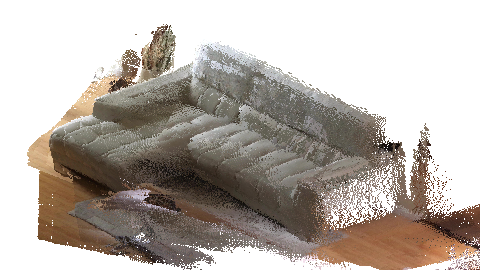}
\qualrow{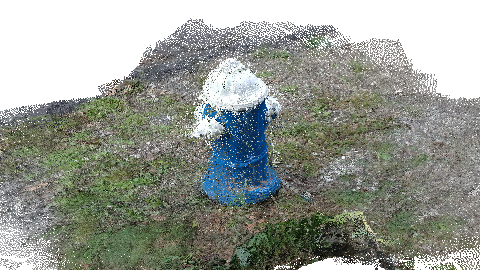}
\qualrow{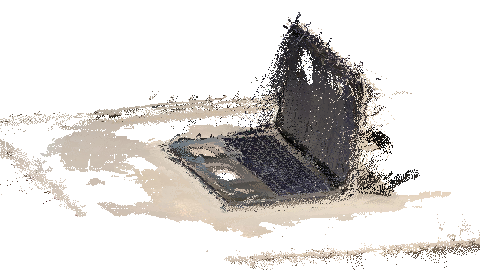}
\qualrow{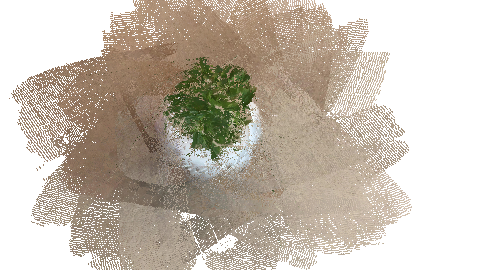}
\qualrow{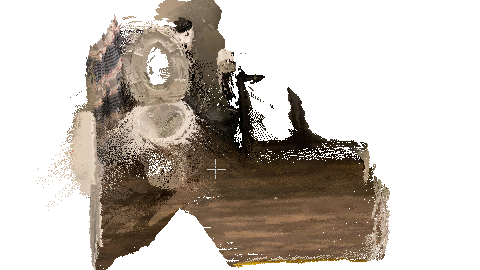}
\qualrow{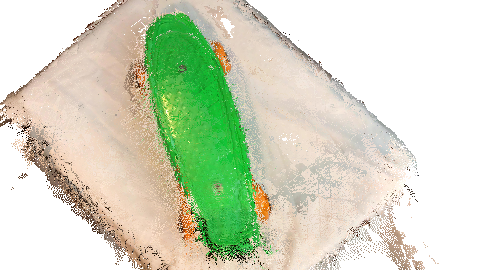}
\qualrow{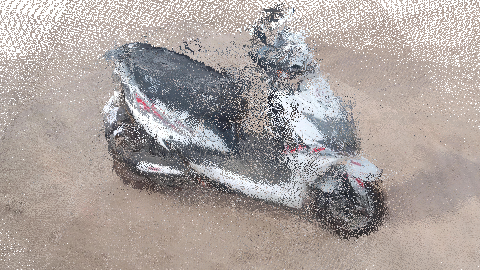}
\qualrow{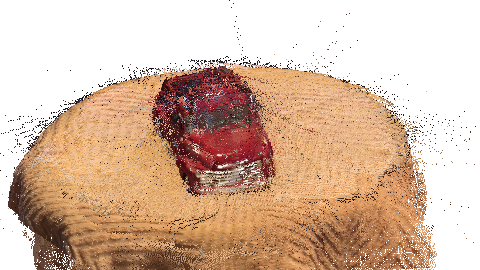}
\qualrow{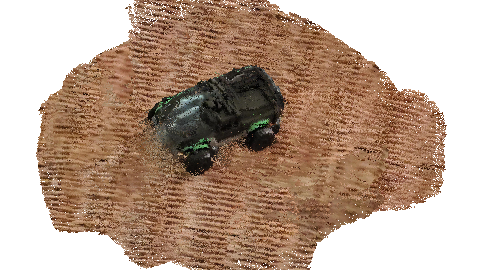}
\qualrow{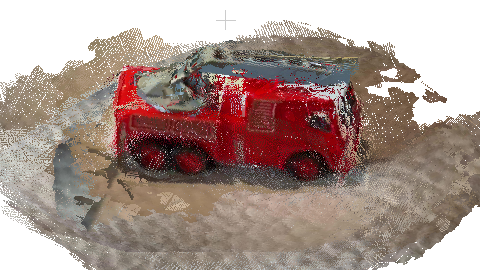}

\section{Failure Cases}
\label{supp:fail}

\begin{figure}[h]
  \centering
  \includegraphics[width=0.49\linewidth]{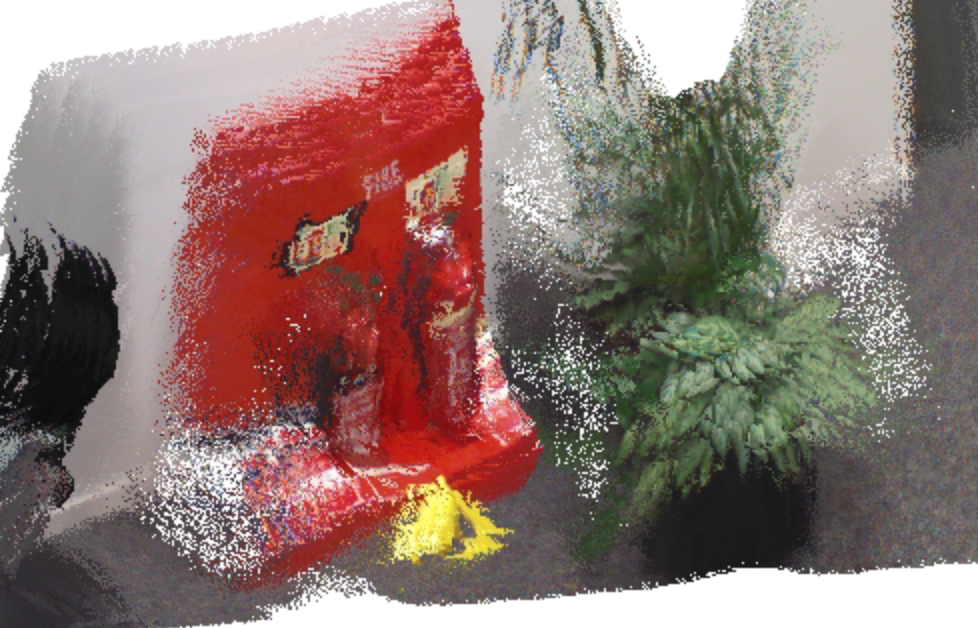}\hfill
  \includegraphics[width=0.49\linewidth]{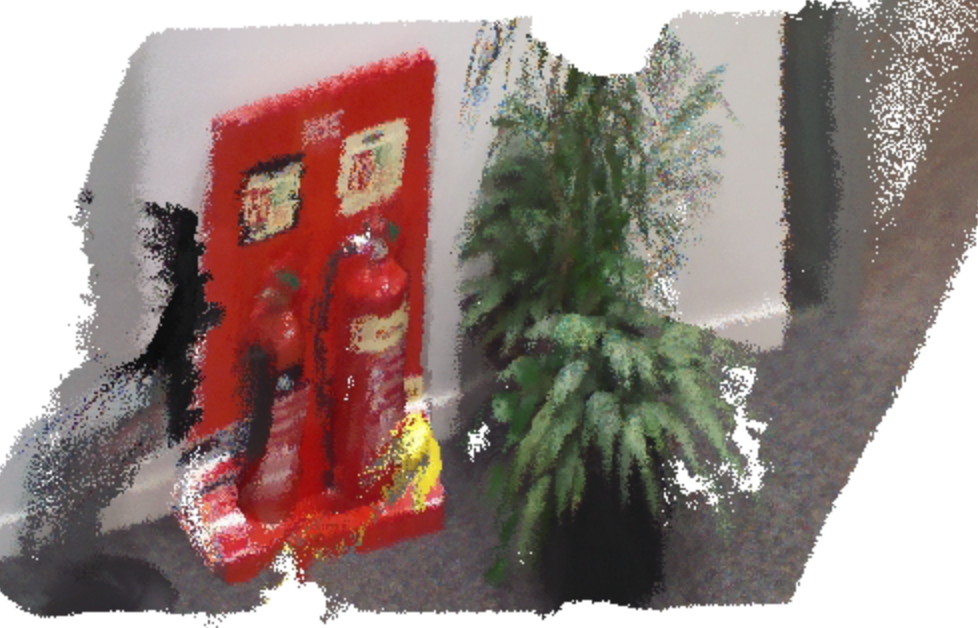}\\[6pt]
  \includegraphics[width=0.49\linewidth]{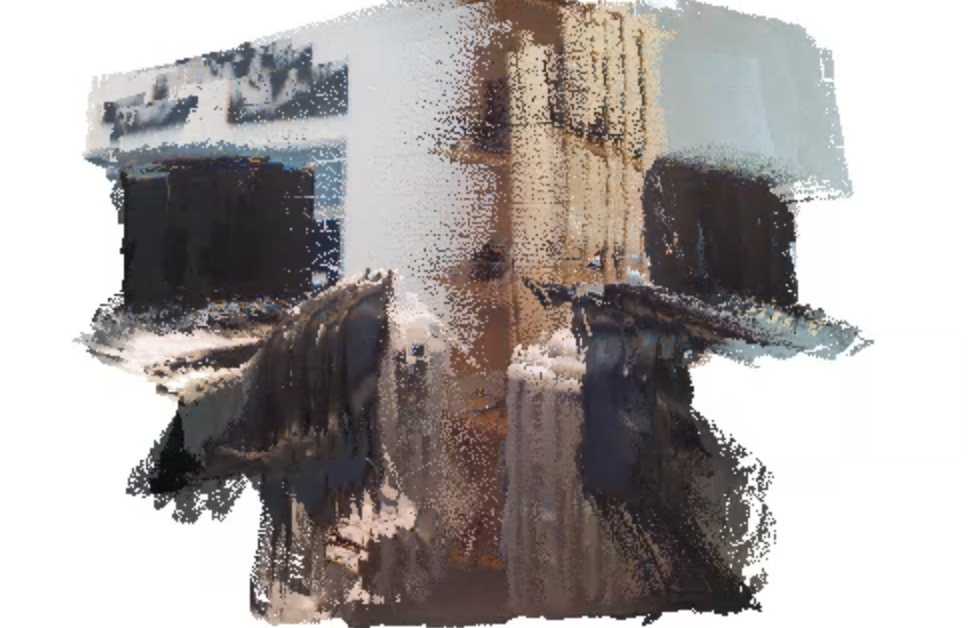}\hfill
  \includegraphics[width=0.49\linewidth]{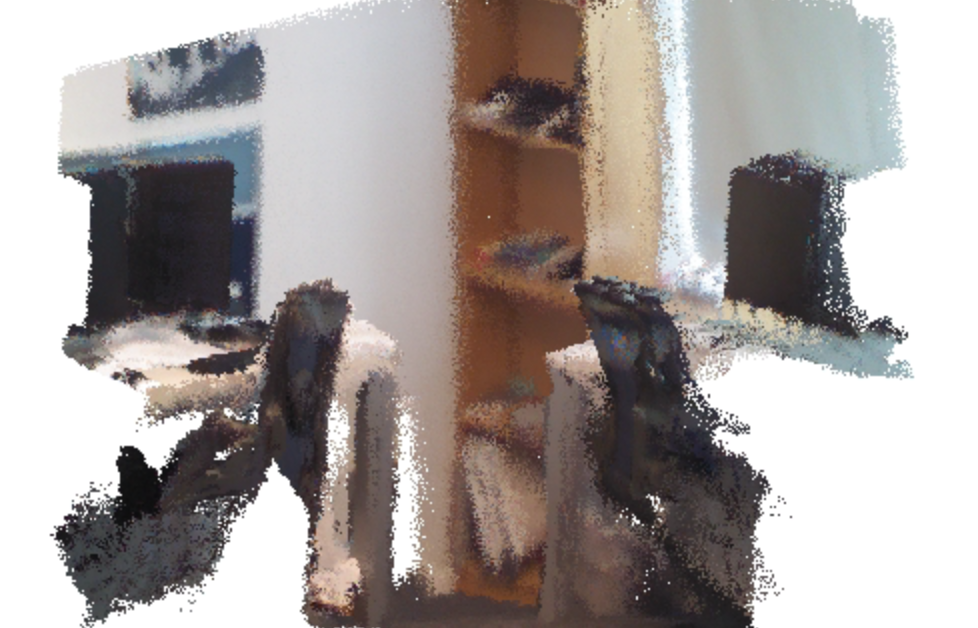}\\[6pt]
  \includegraphics[width=0.49\linewidth]{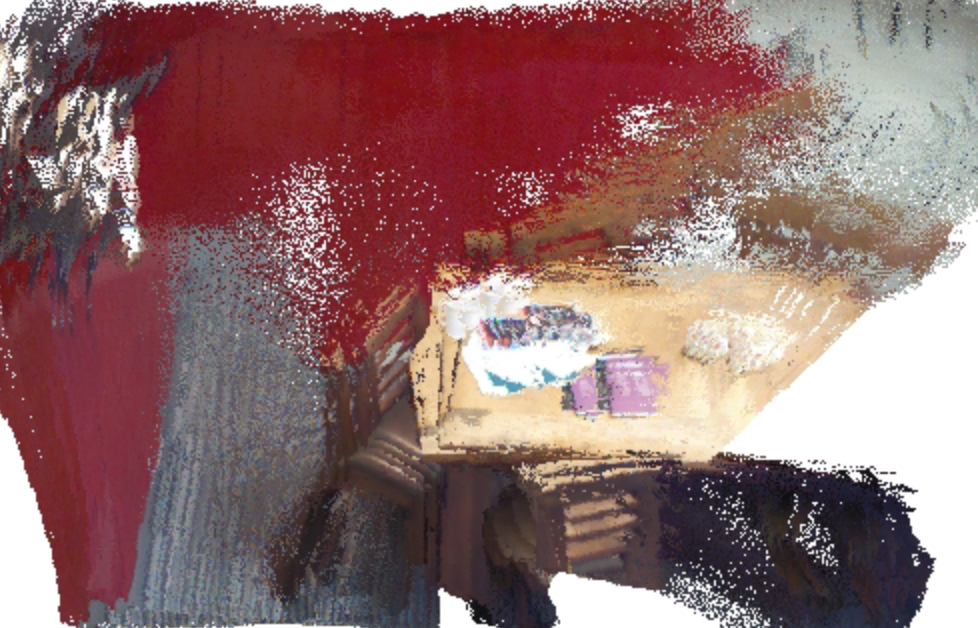}\hfill
  \includegraphics[width=0.49\linewidth]{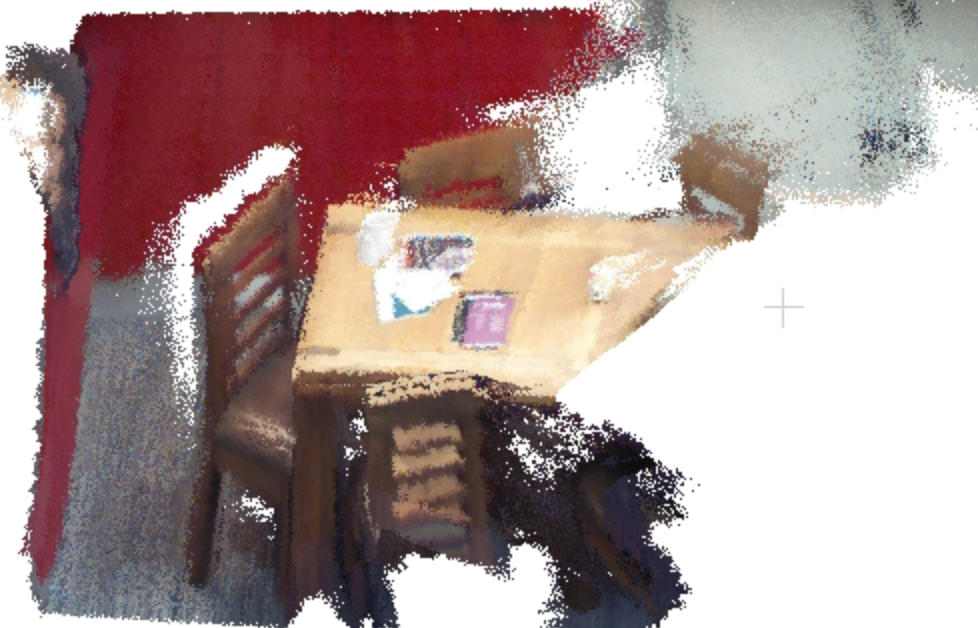}
  \caption{\textbf{Out-of-distribution failures on 7-Scenes.} \emph{Left}: general student, zero-shot. \emph{Right}: domain-specialized student on the same sequence.}
  \label{fig:supp_fail_7s}
\end{figure}

\begin{figure}[h]
  \centering
  \includegraphics[width=0.49\linewidth]{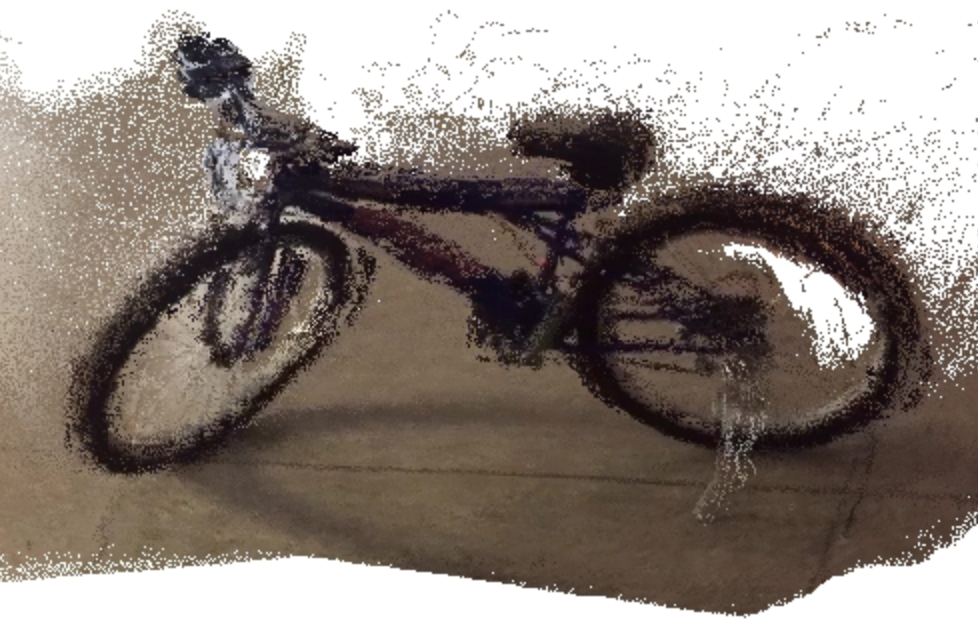}\hfill
  \includegraphics[width=0.49\linewidth]{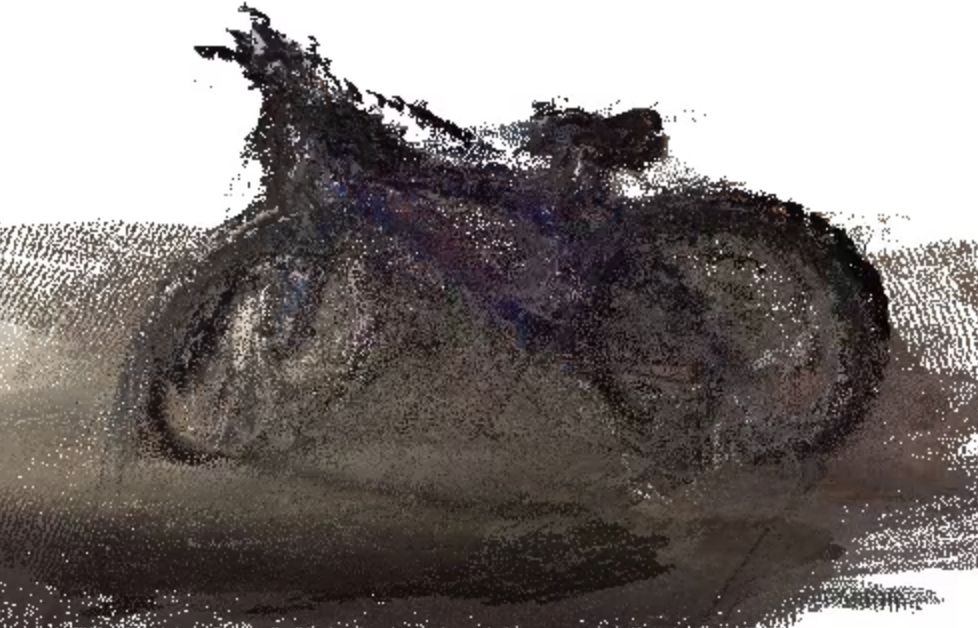}\\[6pt]
  \includegraphics[width=0.49\linewidth]{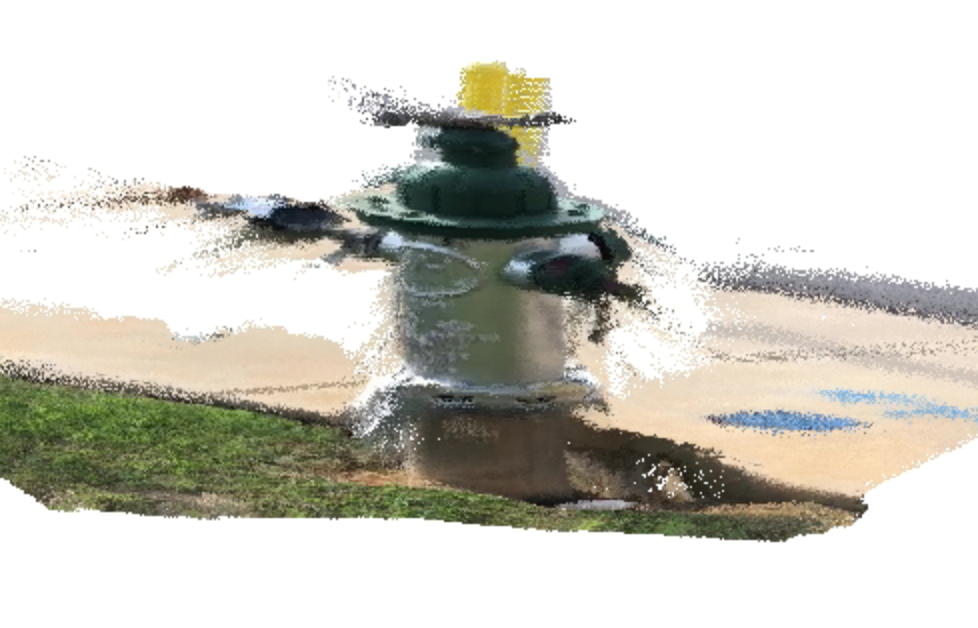}\hfill
  \includegraphics[width=0.49\linewidth]{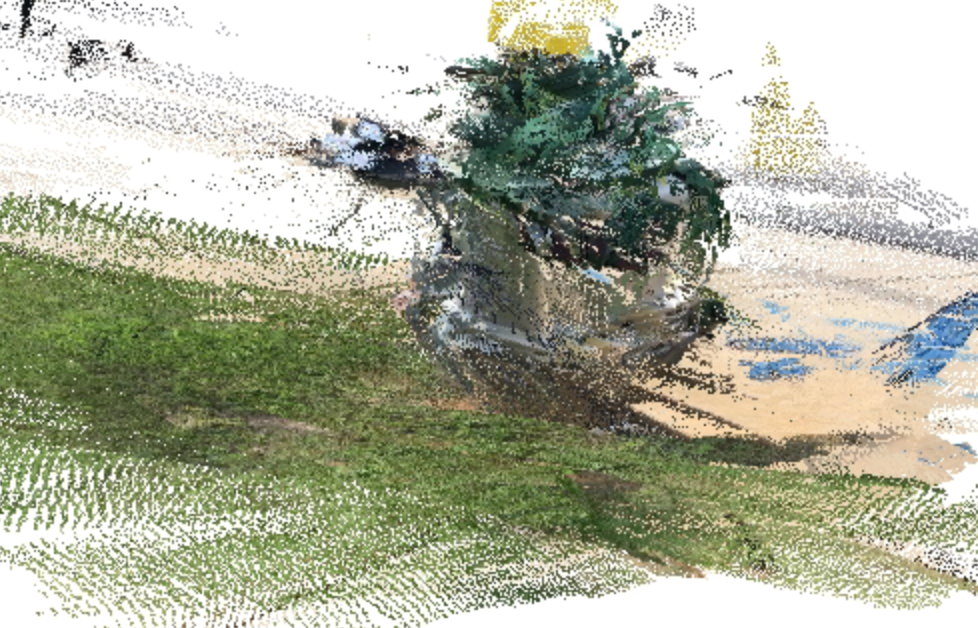}\\[6pt]
  \includegraphics[width=0.49\linewidth]{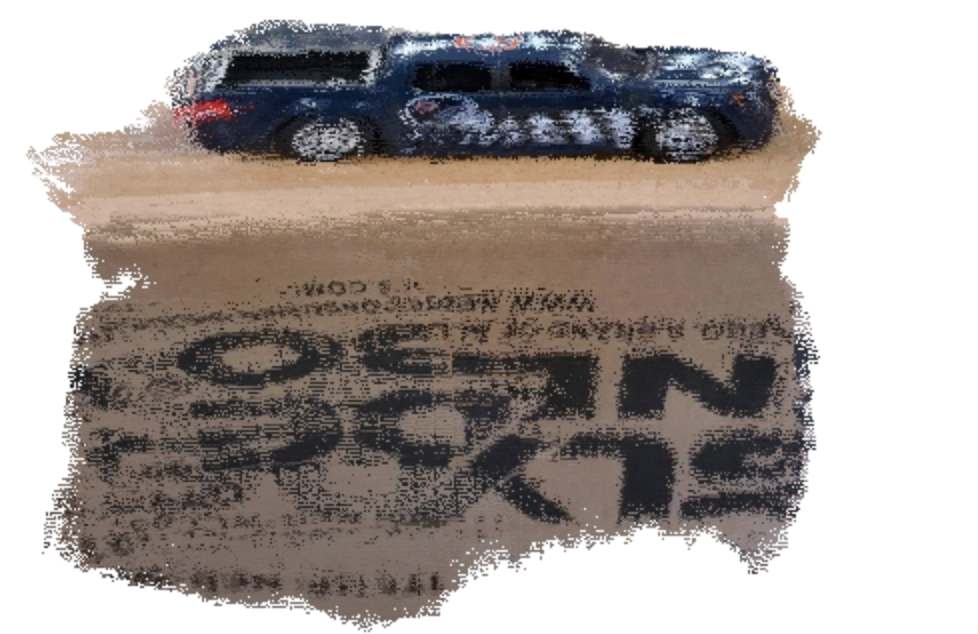}\hfill
  \includegraphics[width=0.49\linewidth]{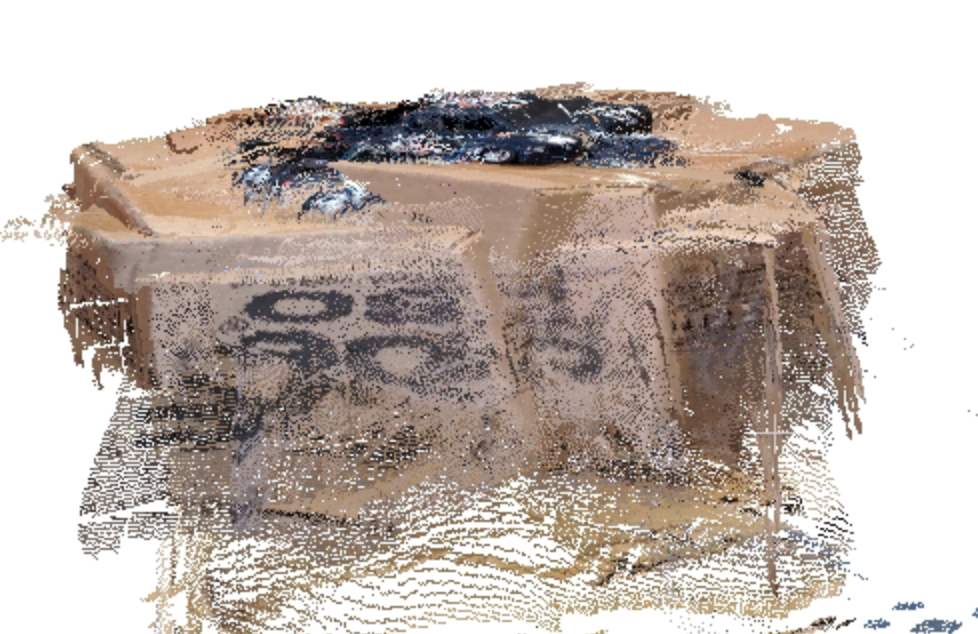}
  \caption{\textbf{Wide-baseline failures on CO3D.} \emph{Left}: temporal (small-baseline) sampling. \emph{Right}: strided (wide-baseline) sampling of the same object, where relative-rotation errors misalign the reconstruction.}
  \label{fig:supp_fail_co3d}
\end{figure}

We show representative failure cases of the student. \cref{fig:supp_fail_7s} compares the general student, evaluated zero-shot, with the domain-specialized student on the same 7-Scenes sequences: without specialization, out-of-distribution pose errors prevent views from correctly aligning (34.84 vs.\ 2.84\,cm completion averaged over 7-Scenes, main Table~1). \cref{fig:supp_fail_co3d} shows CO3D objects reconstructed from temporal and strided samples, where wide-baseline relative-rotation errors misalign the strided reconstruction; front/back confusions on near-symmetric objects, which the teacher shares, dominate this error tail (main Table~2).

\end{document}